\documentclass{article}

\usepackage[preprint,nonatbib]{neurips_2024}

\usepackage[utf8]{inputenc} 
\usepackage[T1]{fontenc}    
\usepackage{hyperref}       
\usepackage{url}            
\usepackage{booktabs}       
\usepackage{amsfonts}       
\usepackage{nicefrac}       
\usepackage{microtype}      
\usepackage[table]{xcolor}         

\definecolor{lightgray}{rgb}{0.94, 0.94, 0.94} 

\usepackage{graphicx}
\usepackage{tablefootnote}
\usepackage{subcaption}
\usepackage{multirow}
\usepackage{amsmath}

\title{Lumen: Unleashing Versatile Vision-Centric Capabilities of Large Multimodal Models}

\author{%
  Yang Jiao$^{1,2,3}$\thanks{This work was done when Yang Jiao was a research intern with Meituan.}\quad Shaoxiang Chen$^{3}$ \quad Zequn Jie$^{3}$ \quad Jingjing Chen$^{1,2}$ \quad Lin Ma$^{3}$ \quad Yu-Gang Jiang$^{1,2}$ \\
  $^{1}$ Shanghai Key Lab of Intell. Info. Processing, School of CS, Fudan University\\
  $^{2}$ Shanghai Collaborative Innovation Center on Intelligent Visual Computing \\
  $^{3}$ Meituan
}

\newcommand{\gray}[1]{\textcolor{gray}{#1}}
\newcommand{\darkgray}[1]{\textcolor{black!65}{#1}}

\begin{document}

\maketitle

\begin{abstract}
  Large Multimodal Model (LMM) is a hot research topic in the computer vision area and has also demonstrated remarkable potential across multiple disciplinary fields. A recent trend is to further extend and enhance the perception capabilities of LMMs. The current methods follow the paradigm of adapting the visual task outputs to language-oriented formats. This adaptation leads to the convenient development of such LMMs with minimal modifications, however, it overlooks the intrinsic characteristics of diverse visual tasks and hinders the learning of perception capabilities. 
   To address this issue, we propose a novel LMM architecture named \textbf{Lumen},
   which decouples the learning of perception capabilities into task-agnostic and task-specific stages. Firstly, Lumen promotes fine-grained vision-language concept alignment, which is the fundamental capability for various visual tasks. Thus the output of the task-agnostic stage is a shared representation for all vision-centric tasks we address in this paper. Afterward, the task-specific decoding is carried out by flexibly routing the shared representation to lightweight task decoders with negligible training efforts. 
   Comprehensive experimental results on a series of vision-centric and VQA benchmarks indicate that our Lumen model not only achieves or surpasses the performance of existing LMM-based approaches in a range of vision-centric tasks while maintaining general visual understanding and instruction following capabilities.
\end{abstract}

\section{Introduction}
\label{sec:intro}
As the Large Language Models (LLMs) currently kindle the spark of Artificial General Intelligence (AGI), Large Multimodal Models (LMMs)~\cite{zhu2023minigpt,liu2023visual} take a step forward by integrating visual modalities with the linguistic prowess of LLMs. With the instruction-following and content-reasoning capabilities inherited from LLMs, the LMM has successfully functioned as a versatile assistant across a wide range of tasks, including visual question answering~\cite{antol2015vqa,qian2023nuscenes}, image captioning~\cite{vaswani2017attention,jiao2022more}, visual commonsense reasoning~\cite{bagherinezhad2016elephants,norlund2021transferring}, etc.

In pursuit of more convenient and efficient human-AI interaction, it is crucial to further explore fundamental vision-centric capabilities encapsulated in the LMMs, which aid in detailed object referencing and dialogue responses. Early works, e.g., MiniGPT-v2~\cite{chen2023minigpt}, Kosmos-2~\cite{peng2023kosmos} and Qwen-VL~\cite{bai2023qwen}, equip the LMM with the grounding ability by reformulating bounding boxes as a sequence of coordinate tokens and adapting them to the language model's output space. Griffon~\cite{zhan2023griffon} extends this design to object detection by meticulously curating a language-prompted detection dataset. Owing to such language model-oriented reformulation, these methods can be implemented with minimal modifications to existing LMMs.

\begin{figure}[tb]
  \centering
  \includegraphics[width=0.9\linewidth]{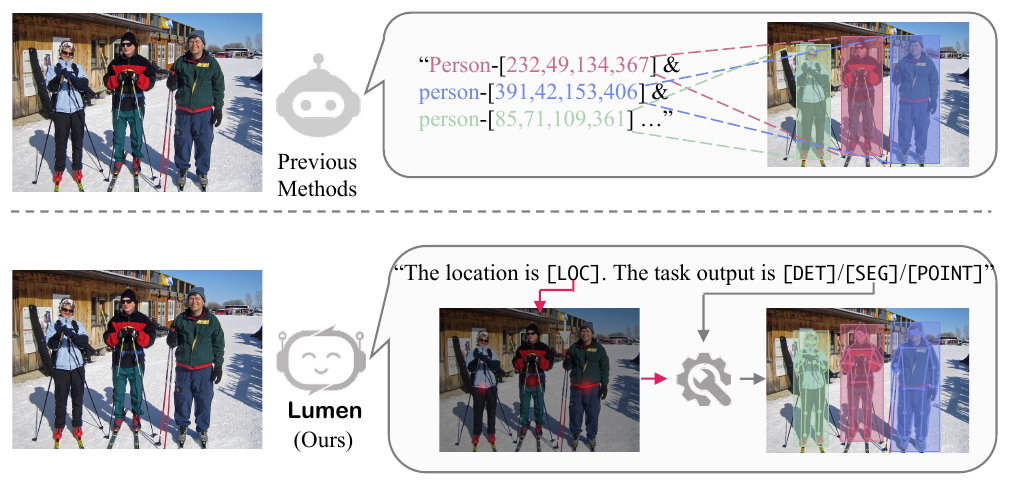}
  \caption{\textbf{Comparison of our proposed Lumen with previous methods.} 
  Our Lumen first predicts unified heatmaps for various tasks. These heatmaps are further used for guiding simple decoding tools with the parsed task-type indicators to support versatile visual tasks. 
  We omit the user instruction of referring to all persons in the image for conciseness. 
  }
  \label{fig:fig1}
\end{figure}

Although convenient, the aforementioned approaches encounter challenges when scaling up to more intricate scenarios and vision-centric tasks. First, those LLMs highly relies on language models's auto-regressive sequence generation method, which leads to high uncertainty when multiple objects are concurrently addressed.
As shown in the first row of Fig~\ref{fig:fig1}, the three persons highlighted with bounding boxes of different colors  lack an inherent order. And imposing a sequential order on them would exacerbate the confusion in the decoding process, as the model would be compelled to produce drastically different outputs after the same word ``person''. 
Although sequence augmentation techniques introduced in Pix2Seq~\cite{chen2021pix2seq} can alleviate this issue, they are tailored for the object detection task and can not be seamlessly transferred to other vision-centric tasks such as instance segmentation and pose estimation.
On the other hand, in contrast to tasks within the NLP field, which often exhibit stronger inter-task correlations~\cite{wei2021finetuned}, vision-centric tasks are inherently discrete due to 
the inductive bias introduced by their distinct task definitions~\cite{xie2023towards}. Therefore, reformulating heterogeneous vision-centric task outputs into language-oriented formats tends to overemphasize the format consistency with language models, while the learning of the underlying visual perception capabilities and intrinsic characteristics of diverse vision tasks is overlooked.

When delving into fundamental vision-centric tasks, namely object detection, instance segmentation, and pose estimation\footnote{We only investigate these three visual tasks in this paper, as LMMs inherently lack image generation capabilities.}, 
we observe that they can be decoupled into task-agnostic and task-specific learning processes. Essentially, these tasks share a common task-agnostic objective of identifying individual instances with an instruction like \emph{``finding the area addressed in instructions''}, while their task-specific definitions introduce different decoding rules for generating diverse formats of outputs (e.g., boxes, masks, or points). 
Compared with the general task-agnostic objective, task-specific outputs relies less on semantics but is more difficult for the LMMs to learn to output.
Based on the above analysis, in this paper, we propose \textbf{Lumen}, a \textbf{L}arge m\textbf{u}ltimodal \textbf{m}odel with vision-centric capabilities \textbf{en}hancement, which decouples the task-agnostic and task-specific learning in two consecutive stages as shown in the Fig~\ref{fig:fig1}. Concretely, in the first stage, the aforementioned visual tasks are unified by reformulating them into the same matching problem. We start by feeding the user's instruction and image into a LMM for content comprehension. The obtained responses contain a designed special token (i.e, \texttt{[LOC]} in Fig~\ref{fig:fig1}) that encapsulates the visual concepts conveyed in the provided instruction, regardless of the specific task. Subsequently, this special token interacts with image patches via a transformer-based aligner to generate a heatmap, wherein the response at each location indicates the matching probability between the instruction and the corresponding image region.
In the second stage, utilizing this heatmap as indicative guidance, task-specific decoding processes are further managed by flexibly assembling predefined decoding rules and lightweight decoders to generate the final outputs with different formats. Due to such decoupled learning behaviors, on the one hand, Lumen can concentrate on promoting fine-grained multimodal content comprehension, rather than being trapped in learning diverse specialized decoding rules lacking in semantics. On the other hand, Lumen can be affected less by the various inductive biases associated with vision-centric tasks within the LLM token space, thereby seamlessly maintaining general visual understanding and instruction following capabilities, as demonstrated in Table~\ref{tab:vqa}.


In summary, our contributions are three folds: (1) We introduce \textbf{Lumen}, a \textbf{L}arge m\textbf{u}ltimodal \textbf{m}odel with vision-centric capabilities \textbf{en}hancement, which unleashes the vision-centric potential of the LMM by decoupling the task-agnostic and task-specific learning processes; (2) Our Lumen can seamlessly adapt to tasks such as object detection, instance segmentation, and pose estimation without requiring specialized dialogue datasets as done in the previous work~\cite{zhan2023griffon}. 
(3) Our Lumen not only matches or exceeds the performances of existing LMM-based approaches on a series of vision-centric tasks, but also maintains general visual understanding and instruction following capabilities. 
\section{Related Work}
\subsection{Large Multimodal Models (LMMs)}
Benefiting from the remarkable reasoning capabilities of Large Language Models (LLMs), LMMs transfer these abilities to the vision domain by aligning visual tokens with LLMs' input space. To achieve this, pioneering work, Flamingo~\cite{alayrac2022flamingo} resamples the visual features and feeds them into attention-based adapter layers inserted in the LLM. Aiming at more comprehensive vision-language alignment, BLIP-2~\cite{li2023blip} designs a Q-Former and jointly performs cross-modal representation learning and generative learning. Inspired by the instruction tuning technique~\cite{wei2021finetuned,ouyang2022training} in NLP field, Instruct-BLIP~\cite{dai2023instructblip}, LLaVA~\cite{liu2023visual} and Mini-GPT4~\cite{zhu2023minigpt} curate high-quality multi-modal instruction data for enhanced instruction-following abilities. However, these methods focus on high-level visual content comprehension and reasoning, ignoring the fundamental visual perception functions, such as object detection, instance segmentation, pose estimation, etc. 

\subsection{Vision Generalist Models}
Generalist models in the vision domain aim at unifying a wide range of vision tasks using one model. Motivated by the success of sequence-to-sequence models in NLP field~\cite{vaswani2017attention}, OFA~\cite{wang2022ofa} and GIT~\cite{wang2022git} unify various tasks in the sequence generation format. Following this trend, Unified-IO~\cite{lu2022unified}, Pix2Seq v2~\cite{chen2022unified} and UniTab~\cite{yang2022unitab} add discrete coordinate tokens into the vocabulary, so as to accommodating more tasks. Moreover, Gato~\cite{reed2022generalist} successfully unifies reinforcement learning tasks as the sequence generation format. Nevertheless, the sequence generation modeling can lead to low inference speeds and degraded performances. Toward this end, Uni-Perceivers~\cite{zhu2022uni,li2023uni} unify different tasks as the maximum likelihood matching problem by calculating representation similarities of all targets and each input. With such an objective, generative and non-generative tasks can be unified by selecting corresponding input and target candidates. However, these generalist models are restricted to pre-defined tasks, failing to be flexibly instructed by natural languages like LMMs. 

\subsection{LMMs with Vision-Centric Capabilities}
To endow LMMs with vision-centric capabilities, two research directions are investigated. On the one hand, a line of work regards LLMs/LMMs as intelligent planners, and allows them to trigger a wide range of task-specific Application Program Interfaces (APIs) according to user's instructions. HuggingGPT~\cite{shen2023hugginggpt} connect GPT with a suite of visual experts. AutoGPT~\cite{yang2023auto} can further execute post-processing programs after detection. Moreover, BuboGPT~\cite{zhao2023bubogpt} further combines visual grounding specialists with LMMs. On the other hand, an alternative approach explores the intrinsic localization potential of LMMs with task-specific modifications. Kosmos-2~\cite{peng2023kosmos}, MiniGPT-v2~\cite{chen2023minigpt}, Qwen-VL~\cite{bai2023qwen} and Shikra~\cite{chen2023shikra} enlarge the vocabulary size of LMMs with discrete coordinate tokens to deal with the visual grounding task~\cite{jiao2023suspected}. LISA~\cite{lai2023lisa} merges the LMM with SAM~\cite{kirillov2023segment} for enhanced reasoning capability in the referring image segmentation scenario~\cite{jiao2021two}. However, these methods do not address fundamental vision tasks like object detection and instance segmentation, where multiple objects should be detected or segmented simultaneously. To address this issue, VisionLLM~\cite{wang2023visionllm} regards an LLM as a DETR-like task decoder and customizes structural prompts for the detection task with Hungarian matching~\cite{kuhn1955hungarian} for label assignment. Griffon~\cite{zhan2023griffon} takes a different approach by leveraging the inherent detection capabilities of the LMM, introducing a language-prompted detection dataset for the instruction tuning. However, these methods leverage discrete coordinate tokens as outputs for different vision tasks, while ignoring their inherent disparities. In this paper, we disentangle the task-agnostic and task-specific learning of various vision-centric tasks to unlock the visual potential of LMMs.

\begin{figure}[tb]
  \centering
  \includegraphics[width=0.9\linewidth]{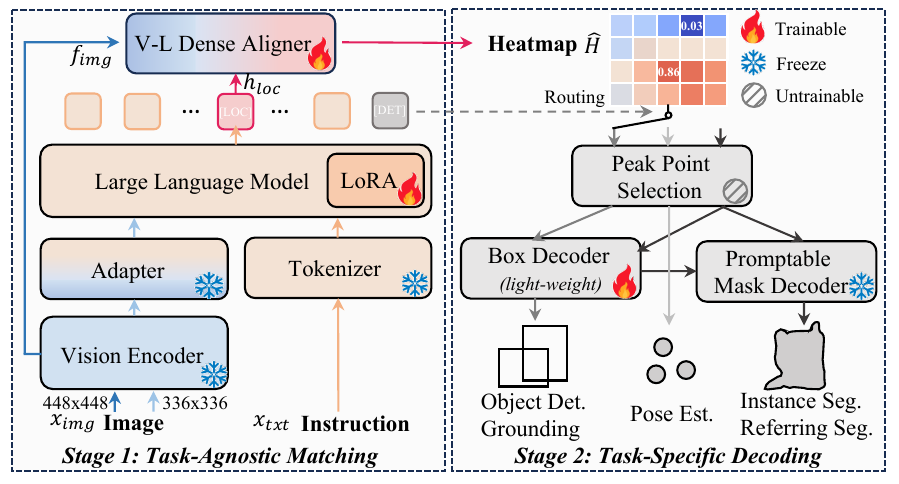}
  \caption{\textbf{Overall framework of the proposed Lumen}. Our Lumen consists of two stages. In the first stage, the input image and the instruction with designed special tokens are embedded and fed into a large language model to interact and comprehend visual and textual contents. Then, the \texttt{[LOC]} token output and high-resolution image features are further aligned to produce a heatmap denoting cross-modal matching probabilities. In the second stage, the heatmap can serve as a strong indication for various vision tasks, and the task outputs can be obtained with lightweight task-specific decoders. The routing of the decoding pathway is determined by the task token (e.g., \texttt{[DET]} in image) output generated in the first stage.
  }
  \label{fig:framework}
\end{figure}

\section{Method}
As shown in Fig~\ref{fig:framework}, our proposed Lumen comprises two consecutive stages. 
In the first stage, we concentrate on promoting fine-grained multimodal content comprehension via densely aligning visual regions and language instructions, disregarding the discrepancies between various vision tasks. In the second stage, with the resulting alignment heatmap from the first stage, task-specific decoding is performed with specialized operations or decoders.
In the following parts, we will elaborate on the detailed designs within each stage.
\subsection{Stage 1: Task-Agnostic Matching}
\subsubsection{Conversation Reformulation} 
\begin{figure}[tb]
  \centering
  \includegraphics[width=\linewidth]{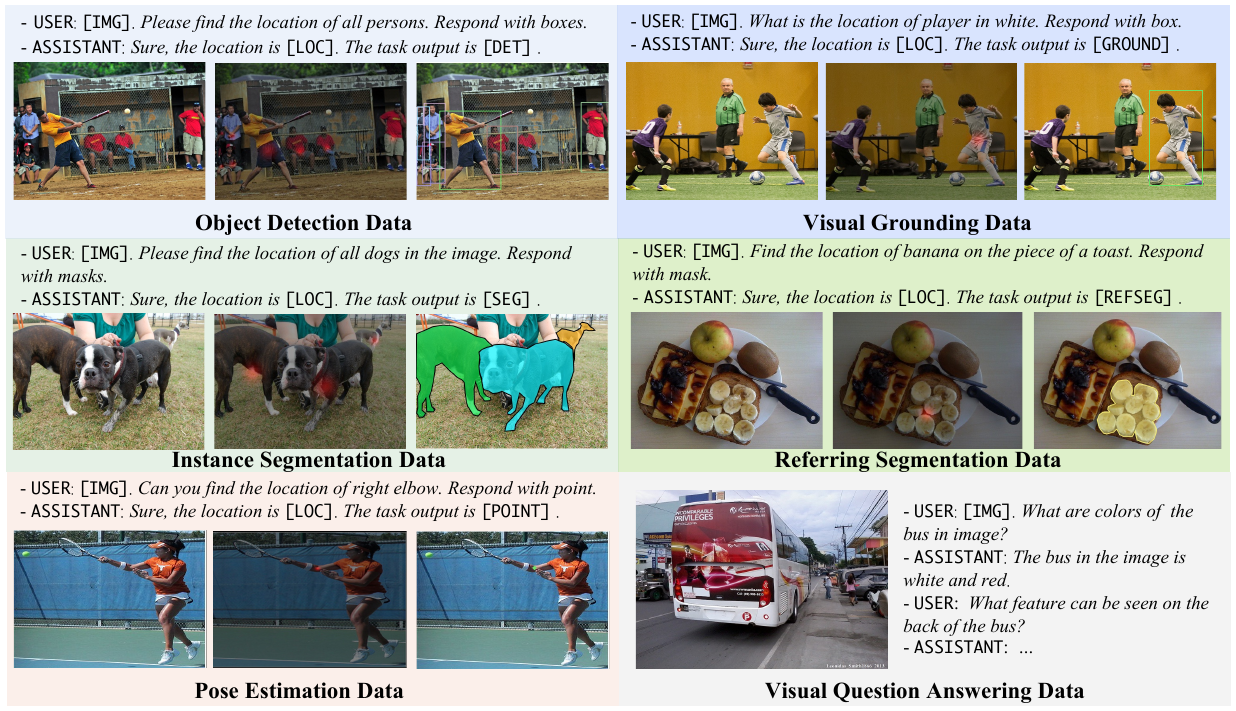}
  \caption{\textbf{The illustration of reformulated conversation data examples of different tasks.} For each reformulated data example, we sequentially present the original image (left), the heatmap generated from annotations (middle), and the task-specific ground truth (right) in the figure.
  }
  \label{fig:conv}
\end{figure}

A preliminary step for adapting vision-centric tasks to LMMs is to reformulate the visual data into conversation formats. For different tasks, we employ a unified instruction-response template: ``\texttt{USER}: \texttt{[IMG]}. \textit{Please find the location of} \texttt{\{description\}}. \textit{Respond with} \texttt{\{format\}}. \texttt{ASSISTANT}: \textit{Sure, the location is} \texttt{[LOC]}. \textit{The task output is} \texttt{[DET]/[SEG]/...}'' Here, \texttt{\{description\}} and \texttt{\{format\}} can be customized according to specific tasks. For vision-centric tasks like object detection, instance segmentation and pose estimation, \texttt{\{description\}} is a certain class name, and \texttt{\{format\}} can be boxes, masks, or points. For vision-language tasks like visual grounding and referring segmentation, \texttt{\{description\}} is the referring sentence and \texttt{\{format\}} can be boxes or masks.
Apart from \texttt{[IMG]} that denotes image features, we also introduce a task-agnostic special token \texttt{[LOC]} and a set of task-specific special tokens, including \texttt{[DET]}, \texttt{[SEG]}, \texttt{[POINT]}, \texttt{[GROUND]} and \texttt{[REFSEG]}. As shown in Fig~\ref{fig:framework}, regardless of task types, \texttt{[LOC]} is tasked to summarize the visual concepts in instructions and further densely align with image regions. Task-specific tokens merely serve as routers for guiding the decoding pathway in the second stage \textbf{without encoding task-specific inductive bias}. Examples of reformulated conversation data of various tasks are illustrated in Fig~\ref{fig:conv}.

\subsubsection{Model Architecture} 
\label{method:arch}
With the given image $x_{img}$ and reformulated instruction $x_{txt}$ as inputs, we first feed them into a conventional LMM (LLaVA-1.5~\cite{liu2023improved} in our implementation) to generate language responses. We also extract high resolution image features $f_{img}\in\mathbb{R}^{32\times32\times256}$ following prior works~\cite{bai2023qwen,zhan2023griffon}:
\begin{equation}
    f_{img} = \mathcal{F_V}(x_{img}),
\end{equation}
where $\mathcal{F_V}$ denotes the vision encoder. Since the adopted vision encoder, CLIP-ViT~\cite{radford2021learning}, is pretrained to accept image inputs of $336\times336$ pixels, we resize its positional embeddings to adapt to higher resolution, e.g., $448\times448$ pixels. Afterward, the feature of \texttt{[LOC]} token $h_{loc}\in\mathbb{R}^{1\times256}$ from the generated response, together with high-resolution image features, are fed into the proposed V-L dense aligner to calculate their matching probabilities, resulting in a heatmap $\hat{H}\in\mathbb{R}^{32\times32}$:
\begin{equation}
    \hat{H} = \mathcal{F_A}(f_{img},h_{loc}),
\end{equation}
where $\mathcal{F_A}$ represents the V-L dense aligner. In our implementation, we employ a lightweight transformer-based architecture as $\mathcal{F_A}$ to interact high resolution image feature tokens and the \texttt{[LOC]} token. More discussions on the architecture design are included in Sec.\ref{sec:exp:comp_anl} and Appendix~\ref{appendix:architecture}.
\subsubsection{Model Training}
To enrich semantics during the dense alignment, we merge data from various tasks by reformulating them into uniform conversation formats as shown in Fig~\ref{fig:conv}. For vision-centric tasks, we hide diverse task-specific output format details and their learning targets are reformulated and unified as heatmaps, where each element denotes the matching probability between the instruction and the corresponding region. To construct the ground-truth heatmap $H$, we use a Gaussian kernel to fill the probabilities into a blank map following prior works~\cite{zhou2019objects,wang2022contextual}:
\begin{equation}
    H_{xy} = \mathrm{exp}(-\frac{(x-l_{x})^{2}+(y-l_{y})^{2}}{2\sigma^{2}}),
\end{equation}
where the $(l_{x},l_{y})$ is the location of the one referred by the instruction, and $\sigma$ is the standard deviation. For object detection and visual grounding, $(l_{x},l_{y})$ is the center coordinate of each object, and $\sigma$ is object size-adaptive following~\cite{zhou2019objects}. For instance segmentation and referring segmentation, we first convert their masks into bounding boxes by calculating enclosing rectangles of masks, and then calculate $(l_{x},l_{y})$ and $\sigma$ with the same rules as detection and visual grounding tasks. For pose estimation, $(l_{x},l_{y})$ is the coordinate of the annotated keypoint, $\sigma$ is a hyper-parameter designated as 2. 

With the ground-truth heatmap $H$ and predicted heatmap $\hat{H}$, we apply Gaussian focal loss following~\cite{zhou2019objects} as:
\begin{equation}
\mathcal{L}_h = -\frac{1}{N} \sum_{xy} \left\{
    \begin{array}{ll}
      (1 - \hat{H}_{xy})^\alpha \log(\hat{H}_{xy}) & \text{if } H_{xy} = 1, \\
      (1 - H_{xy})^\beta (\hat{H}_{xy})^\alpha \log(1 - \hat{H}_{xy}) & \text{otherwise}.
    \end{array}
\right.
\end{equation}
$\alpha$ and $\beta$ are hyper-parameters of the focal loss, and $N$ is the number of the location where the ground-truth matching probability equals to 1. We set $\alpha=2$ and $\beta=4$ in our implementation. For supervising the language outputs, we use the standard cross-entropy loss $\mathcal{L}_{t}$ following previous LMM-based methods~\cite{zhan2023griffon,bai2023qwen}. The overall loss function $\mathcal{L}$ can be formulated as:
\begin{equation}
    \mathcal{L} = \lambda_{h}\mathcal{L}_{h} + \lambda_{t}\mathcal{L}_{t},
\end{equation}
where $\lambda_{h}$ and $\lambda_{t}$ are hyper-parameters for balancing two losses.
During training, since the vision encoder, adapter, tokenizer, and large language model parts in Fig~\ref{fig:framework} inherit the weights of LLaVA, we only finetune the large language model using the LoRA~\cite{hu2021lora} technique and train the V-L dense aligner under the supervision of the loss function $\mathcal{L}$.
\subsection{Stage 2: Task-Specific Decoding}
\label{method:stage2}

\textbf{Decoding Modules.}\quad
We first introduce the operations of the three key modules illustrated in Fig~\ref{fig:framework}. 
(1) \textbf{Peak point selection} parses the heatmap into a set of points, where each point indicates the center location of an identified object or keypoint. Specifically, we filter the locations with heatmap response greater than their 8 surrounding neighbors as candidate peak points, and then retain the top $K$ candidates as the selected results. The value of $K$ can vary across different tasks, which will be elaborated on in Sec.\ref{sec:exp:exp_setup}. 
(2) \textbf{Box decoder} is used for further regressing the extent of the objects designated by the selected peak points. For efficiency, instead of introducing an additional network as the box decoder, we reuse the V-L dense aligner by appending two new learnable special tokens after the \texttt{[LOC]} token as additional text-side inputs. Accordingly, the V-L dense aligner will generate two additional 1-channel map predictions, which are used for regressing the height and width under the supervision of the L1 loss functions similar to~\cite{zhou2019objects}. 
(3) \textbf{Promptable mask decoder} can accept both points and boxes as visual instructions to generate the mask for the referred object. We directly use the SAM model~\cite{kirillov2023segment} for executing the above process without finetuning. 
In general, these decoding modules introduce negligible training costs and can be easily implemented.

\textbf{Decoding Pathways.}\quad
On the basis of these three decoding modules, we customize their different cooperation patterns to perform diverse tasks, which will be introduced as follows: 
(1) \textbf{Pose estimation} requires to return the keypoint coordinates. We can easily obtain these coordinates by parsing the predicted heatmap with the peak point selection operation. 
(2) \textbf{Object detection} and \textbf{visual grounding} share the same output format of bounding box, therefore, we employ the identical decoding pathway for them. Specifically, as shown in Fig~\ref{fig:framework}, we feed the heatmap into the cascaded peak point selection and box decoder modules to generate bounding boxes. 
(3) \textbf{Instance segmentation} and \textbf{referring segmentation} share a relationship analogous to the one between object detection and visual grounding, therefore, we also adopt the same decoding pathway for them. Concretely, we first parse both the center point and bounding box of the object, and then, we use them as visual instructions to guide the promptable mask decoder to generate the mask prediction.





\section{Experiments}
\subsection{Experimental Setup}
\label{sec:exp:exp_setup}

\textbf{Datasets.}\quad
Our training data consists of datasets from the following different tasks. (1) For object detection, we use MSCOCO~\cite{lin2014microsoft}, Objects365~\cite{shao2019objects365} and OpenImages~\cite{kuznetsova2020open}. (2) For visual grounding, we use RefCOCO, RefCCO+ and RefCOCOg~\cite{yu2016modeling}. (3) For pose estimation, we use MSCOCO keypoints~\cite{lin2014microsoft} and AIC~\cite{wu2017ai}. (4) For visual question answering, we employ a subset of ShareGPT4V dataset~\cite{chen2023sharegpt4v} with 665K samples. It is worth noting that for instance segmentation and referring segmentation, since we first transform their masks into bounding boxes as mentioned in Sec~\ref{method:arch}, the constructed heatmaps are the same as the object detection. Additionally, given that we do not finetune the mask decoder in the second decoding stage, we actually do not use segmentation annotations throughout the entire training process. 

\textbf{Model Configurations.}\quad
\label{appendix:model_configurations}
For the first task-agnostic matching stage, we utilize pre-trained CLIP ViT-L/14-336~\cite{radford2021learning} and LLaVA-7B-v1.5~\cite{liu2023visual} as our vision encoder and large multimodal model, respectively.
For the second decoding stage, since the different decoding modules and pathways have been introduced in Sec~\ref{method:stage2}, we primarily specify the $K$ value choices and corresponding post-processing details of different tasks here. (1) For object detection and instance segmentation, which generally includes multiple objects in a given image, we set $K=100$ to generate 100 box candidates. Then we apply regular NMS~\cite{hosang2017learning} to filter redundant boxes, and the remaining boxes are used for further prompt mask decoder to generate instance segmentation results. (2) For visual grounding and referring segmentation, given that only one object is referred by a referring sentence, we set $K=1$ to only generate one predicted box/mask, and no post-processing is required. (3) For pose estimation, we follow previous works~\cite{geng2023instructdiffusion,chen2021pix2seq} to first crop the single object from the image using bounding boxes. Then, for each keypoint category, we set $K=1$ to extract the point with the highest matching probability as the prediction result. 

\textbf{Training Details.}\quad
\label{appendix:training_details}
For the task-agnostic stage, our training comprises two phases. (1) In Phase 1, we mix the object detection, visual grounding and pose estimation data with sampling rates of 0.69, 0.23, 0.08, respectively, for balanced data sampling.
(2) In Phase 2, we mix the visual question-answering, object detection, visual grounding and pose estimation data with sample rates of 0.67, 0.23, 0.07, 0.03, respectively. We set the batch size to 160 and train the first step for 50,000 steps and the second step for 10,000 steps on 8 NVIDIA 80G A100 GPUs. 
The loss function balance terms $\lambda_h$ and $\lambda_t$ are both set to 1. For each phase, we use AdamW as the optimizer with an initial learning rate of $3\times10^{-4}$ and weight decay of 0. During training, we do not calculate heatmap loss $\mathcal{L}_{h}$ for visual question-answering data. We do not use any data augmentation techniques for all tasks.

\textbf{Evaluation Metrics.}\quad
We adopt evaluation metrics commonly used within each field of task. For object detection and instance segmentation, we use mAP based on box IoU and mask IoU, respectively. For pose estimation, we use mAP based on OKS (object keypoint similarity). For visual question-answering, we comply with the evaluation protocol of each individual benchmark.

\begin{table*}[!t]
\centering
\caption{\textbf{Results on fundamental vision-centric tasks and vision-language tasks.} 
We use ``-'' and gray cell to indicate the result is not reported and the corresponding task is not supported by the method, respectively.}
\label{tab:ver}
\scalebox{0.8}{
\begin{tabular}{lccccccccccc}
\toprule
\multicolumn{1}{l|}{Method}           & \multicolumn{3}{c|}{Object Det.}                                   & \multicolumn{3}{c|}{Instance Seg.}                                 & \multicolumn{3}{c|}{Pose Est.}                                     & \multicolumn{1}{c|}{Grounding}     & Refer Seg. \\ \cline{2-12} 
\multicolumn{1}{l|}{}                 & AP            & AP$_{50}$          & \multicolumn{1}{c|}{AP$_{75}$}          & AP            & AP$_{50}$          & \multicolumn{1}{c|}{AP$_{75}$}          & AP            & AP$_{50}$          & \multicolumn{1}{c|}{AP$_{75}$}          & \multicolumn{1}{c|}{AP$_{50}$}          & cIoU           \\ \midrule
\multicolumn{12}{c}{\gray{\textit{\textbf{Task-specific Specialists}}}}                                                                                                                                                                                                                                                         \\ \midrule
\multicolumn{1}{l|}{Faster R-CNN~\cite{ren2015faster}}  & \gray{40.3}          & \gray{61.0}          & \multicolumn{1}{c|}{\gray{44.0}}          & \cellcolor{lightgray}              & \cellcolor{lightgray}             & \multicolumn{1}{c|}{\cellcolor{lightgray}}             & \cellcolor{lightgray}             & \cellcolor{lightgray}             & \multicolumn{1}{c|}{\cellcolor{lightgray}}             & \multicolumn{1}{c|}{\cellcolor{lightgray}}             & \cellcolor{lightgray}              \\
\multicolumn{1}{l|}{DETR~\cite{carion2020end}}             & \gray{43.3}          & \gray{63.1}          & \multicolumn{1}{c|}{\gray{45.9}}          & \cellcolor{lightgray}             & \cellcolor{lightgray}             & \multicolumn{1}{c|}{\cellcolor{lightgray}}             & \cellcolor{lightgray}             & \cellcolor{lightgray}             & \multicolumn{1}{c|}{\cellcolor{lightgray}}             & \multicolumn{1}{c|}{\cellcolor{lightgray}}             & \cellcolor{lightgray}              \\
\multicolumn{1}{l|}{Mask R-CNN~\cite{he2017mask}}       & \gray{41.0}          & \gray{61.7}          & \multicolumn{1}{c|}{\gray{44.9}}          & \gray{37.1}          & \gray{58.4}          & \multicolumn{1}{c|}{\gray{40.1}}          & \cellcolor{lightgray}             & \cellcolor{lightgray}             & \multicolumn{1}{c|}{\cellcolor{lightgray}}             & \multicolumn{1}{c|}{\cellcolor{lightgray}}             & \cellcolor{lightgray}             \\
\multicolumn{1}{l|}{PolarMask~\cite{xie2020polarmask}}       & \cellcolor{lightgray}             & \cellcolor{lightgray}             & \multicolumn{1}{c|}{\cellcolor{lightgray}}             & \gray{30.5}          & \gray{52.0}          & \multicolumn{1}{c|}{\gray{31.1}}          & \cellcolor{lightgray}             & \cellcolor{lightgray}             & \multicolumn{1}{c|}{\cellcolor{lightgray}}             & \multicolumn{1}{c|}{\cellcolor{lightgray}}             & \cellcolor{lightgray}              \\
\multicolumn{1}{l|}{CPM~\cite{xie2020polarmask}}              & \cellcolor{lightgray}             & \cellcolor{lightgray}             & \multicolumn{1}{c|}{\cellcolor{lightgray}}             & \cellcolor{lightgray}             & \cellcolor{lightgray}             & \multicolumn{1}{c|}{\cellcolor{lightgray}}             & \gray{62.7}          & \gray{86.2}          & \multicolumn{1}{c|}{\gray{70.9}}          & \multicolumn{1}{c|}{\cellcolor{lightgray}}             & \cellcolor{lightgray}              \\
\multicolumn{1}{l|}{RTMPose~\cite{jiang2023rtmpose}}          & \cellcolor{lightgray}             & \cellcolor{lightgray}             & \multicolumn{1}{c|}{\cellcolor{lightgray}}             & \cellcolor{lightgray}             & \cellcolor{lightgray}             & \multicolumn{1}{c|}{\cellcolor{lightgray}}             & \gray{68.2}          & \gray{88.3}          & \multicolumn{1}{c|}{\gray{75.9}}          & \multicolumn{1}{c|}{\cellcolor{lightgray}}             & \cellcolor{lightgray}              \\
\multicolumn{1}{l|}{MDETR~\cite{kamath2021mdetr}}            & \cellcolor{lightgray}             & \cellcolor{lightgray}             & \multicolumn{1}{c|}{\cellcolor{lightgray}}             & \cellcolor{lightgray}             & \cellcolor{lightgray}             & \multicolumn{1}{c|}{\cellcolor{lightgray}}             & \cellcolor{lightgray}             & \cellcolor{lightgray}             & \multicolumn{1}{c|}{\cellcolor{lightgray}}             & \multicolumn{1}{c|}{\gray{83.4}}          & \cellcolor{lightgray}              \\
\multicolumn{1}{l|}{SeqTR~\cite{zhu2022seqtr}}            & \cellcolor{lightgray}            & \cellcolor{lightgray}            & \multicolumn{1}{c|}{\cellcolor{lightgray}}             & \cellcolor{lightgray}             & \cellcolor{lightgray}             & \multicolumn{1}{c|}{\cellcolor{lightgray}}             & \cellcolor{lightgray}             & \cellcolor{lightgray}             & \multicolumn{1}{c|}{\cellcolor{lightgray}}             & \multicolumn{1}{c|}{\gray{82.7}}          & \gray{64.7}           \\
\multicolumn{1}{l|}{LAVT~\cite{yang2022lavt}}             & \cellcolor{lightgray}             & \cellcolor{lightgray}             & \multicolumn{1}{c|}{\cellcolor{lightgray}}             & \cellcolor{lightgray}             & \cellcolor{lightgray}             & \multicolumn{1}{c|}{\cellcolor{lightgray}}             & \cellcolor{lightgray}             & \cellcolor{lightgray}             & \multicolumn{1}{c|}{\cellcolor{lightgray}}             & \multicolumn{1}{c|}{\cellcolor{lightgray}}             & \gray{61.2}           \\
\multicolumn{1}{l|}{ReLA~\cite{liu2023gres}}             & \cellcolor{lightgray}             & \cellcolor{lightgray}             & \multicolumn{1}{c|}{\cellcolor{lightgray}}             & \cellcolor{lightgray}             & \cellcolor{lightgray}             & \multicolumn{1}{c|}{\cellcolor{lightgray}}             & \cellcolor{lightgray}             & \cellcolor{lightgray}             & \multicolumn{1}{c|}{\cellcolor{lightgray}}             & \multicolumn{1}{c|}{\cellcolor{lightgray}}             & \gray{65.0}           \\ \midrule
\multicolumn{12}{c}{\darkgray{\textit{\textbf{Vision Generalists}}}}                                                                                                                                                                                                                                                  \\ \bottomrule
\multicolumn{1}{l|}{Pix2Seq-v2~\cite{chen2021pix2seq}}       & \darkgray{46.5}          & \darkgray{-}             & \multicolumn{1}{c|}{\darkgray{-}}             & \darkgray{38.2}          & \darkgray{-}             & \multicolumn{1}{c|}{\darkgray{-}}             & \darkgray{64.8}             & \darkgray{-}             & \multicolumn{1}{c|}{\darkgray{-}}             & \multicolumn{1}{c|}{\cellcolor{lightgray}}             &  \cellcolor{lightgray}              \\
\multicolumn{1}{l|}{Uni-Perceiver-v2~\cite{li2023uni}} & \darkgray{58.6}          & \darkgray{-}             & \multicolumn{1}{c|}{\darkgray{-}}             & \darkgray{50.6}          & \darkgray{-}             & \multicolumn{1}{c|}{\darkgray{-}}             & \cellcolor{lightgray}             & \cellcolor{lightgray}             & \multicolumn{1}{c|}{\cellcolor{lightgray}}             & \multicolumn{1}{c|}{\cellcolor{lightgray}}             & \cellcolor{lightgray}              \\
\multicolumn{1}{l|}{mPLUG-2~\cite{xu2023mplug}}          & \darkgray{46.9}          & \darkgray{-}             & \multicolumn{1}{c|}{\darkgray{-}}             & \darkgray{40.6}          & \darkgray{-}             & \multicolumn{1}{c|}{\darkgray{-}}             & \cellcolor{lightgray}             & \cellcolor{lightgray}             & \multicolumn{1}{c|}{\cellcolor{lightgray}}             & \multicolumn{1}{c|}{\darkgray{84.7}}          & \cellcolor{lightgray}              \\
\multicolumn{1}{l|}{VisionLLM~\cite{wang2023visionllm}}        & \darkgray{44.6}          & \darkgray{64.0}          & \multicolumn{1}{c|}{\darkgray{48.1}}          & \darkgray{25.1}          & \darkgray{50.0}          & \multicolumn{1}{c|}{\darkgray{22.4}}          & \cellcolor{lightgray}             & \cellcolor{lightgray}             & \multicolumn{1}{c|}{\cellcolor{lightgray}}             & \multicolumn{1}{c|}{\darkgray{-}}             & \cellcolor{lightgray}              \\ \midrule
\multicolumn{12}{c}{\textit{\textbf{LMM Generalists}}}                                                                                                                                                                                                                                                     \\ \midrule
\multicolumn{1}{l|}{Shikra-7B~\cite{chen2023shikra}}        & \cellcolor{lightgray}             & \cellcolor{lightgray}             & \multicolumn{1}{c|}{\cellcolor{lightgray}}             & \cellcolor{lightgray}             & \cellcolor{lightgray}             & \multicolumn{1}{c|}{\cellcolor{lightgray}}             & \cellcolor{lightgray}             & \cellcolor{lightgray}             & \multicolumn{1}{c|}{\cellcolor{lightgray}}             & \multicolumn{1}{c|}{82.3}          & \cellcolor{lightgray}              \\
\multicolumn{1}{l|}{MiniGPT-v2-7B~\cite{chen2023minigpt}}       & \cellcolor{lightgray}             & \cellcolor{lightgray}             & \multicolumn{1}{c|}{\cellcolor{lightgray}}             & \cellcolor{lightgray}             & \cellcolor{lightgray}             & \multicolumn{1}{c|}{\cellcolor{lightgray}}             & \cellcolor{lightgray}             & \cellcolor{lightgray}             & \multicolumn{1}{c|}{\cellcolor{lightgray}}             & \multicolumn{1}{c|}{\textbf{84.4}} & \cellcolor{lightgray}              \\
\multicolumn{1}{l|}{Griffon-13B~\cite{zhan2023griffon}}      & 23.2          & 37.6          & \multicolumn{1}{c|}{23.4}          & \cellcolor{lightgray}             & \cellcolor{lightgray}             & \multicolumn{1}{c|}{\cellcolor{lightgray}}             & \cellcolor{lightgray}             & \cellcolor{lightgray}             & \multicolumn{1}{c|}{\cellcolor{lightgray}}             & \multicolumn{1}{c|}{81.5}          & \cellcolor{lightgray}              \\
\multicolumn{1}{l|}{InstructCV~\cite{gan2023instructcv}}       & -             & 48.5          & \multicolumn{1}{c|}{-}             & \cellcolor{lightgray}             & \cellcolor{lightgray}             & \multicolumn{1}{c|}{\cellcolor{lightgray}}             & \cellcolor{lightgray}             & \cellcolor{lightgray}             & \multicolumn{1}{c|}{\cellcolor{lightgray}}             & \multicolumn{1}{c|}{\cellcolor{lightgray}}             & \cellcolor{lightgray}              \\
\multicolumn{1}{l|}{\textbf{Lumen-7B (Ours)}}     & \textbf{35.3} & \textbf{53.2} & \multicolumn{1}{c|}{\textbf{35.8}} & \textbf{30.4} & \textbf{49.8} & \multicolumn{1}{c|}{\textbf{31.0}} & \textbf{67.2} & \textbf{90.4} & \multicolumn{1}{c|}{\textbf{75.6}} & \multicolumn{1}{c|}{83.6}          & \textbf{65.1}  \\ \bottomrule
\end{tabular}
}
\end{table*}

\begin{table}[!t]
\centering
\caption{\textbf{Results on prevalent VQA benchmarks.} Here, we employ English MMBench dev, SEEDBench image, MME test, MMMU val and MathVista mini sets for evaluation. }
\label{tab:vqa}
\scalebox{0.8}{
\begin{tabular}{l|c|ccccc}
\toprule
Method       & Param & MMBench & SEED & MME      & MMMU & MathVista \\ \midrule
InstructBLIP~\cite{dai2023instructblip} & 7B    & 36.0    & 58.8      & 1213/292      & 32.9 & \textbf{25.3}      \\
MiniGPT-4~\cite{zhu2023minigpt}    & 7B    & 24.3    & 47.4      & 582/144  & -    & 23.1      \\
Shikra~\cite{chen2023shikra}       & 7B    & 58.8    & -         & -         & -    & -         \\
Qwen-VL-Chat~\cite{bai2023qwen} & 7B    & 60.6    & 58.2      & 1488/\textbf{361}   & \textbf{35.9} & -         \\
LLaVA-v1.5~\cite{liu2023improved}   & 7B   & 64.3    & \textbf{66.1}      & \textbf{1511}/296 & 35.6 & 23.5      \\
\textbf{Lumen (Ours)} & 7B    & \textbf{64.9}    & 65.8      & 1426/332  & 35.2    & 24.6         \\ \bottomrule
\end{tabular}}
\end{table}

\subsection{Results on Versatile Tasks}
\label{sec:exp:main_rst}
We evaluate our method on vision-centric and vision-language tasks as shown in Table~\ref{tab:ver} and~\ref{tab:vqa}. We categorize existing approaches into three groups, namely \textit{``task-specific specialists''}, \textit{``vision generalists''} and \textit{``LMM generalists''}, according to their functions and architectures. 
(1) \textbf{Task-specific specialists} are customized models in different fields. They have diverse architectural designs and are limited to a single task. (2) \textbf{Vision generalists} pursue handling multiple tasks with a unified architecture. To excel in fundamental visual perception tasks, they typically utilize a powerful vision encoder or additional designs (e.g, data augmentations or decoding strategies) adaptive to vision-centric tasks. (3) \textbf{LMM generalists} aim to resolve every task in a conversational format. Focusing on improving the conversational quality, they adopt vision encoders proficient in multimodal content comprehension. In line with these methods, our Lumen pursues \emph{\textbf{unleashing the inherent vision-centric capabilities of LMMs}}.

\textbf{Object Detection \& Instance Segmentation.}\quad
Object detection and instance segmentation require the model to make dense predictions across the image, therefore they pose great challenges in capturing fine-grained object cues. COCO val set is used for evaluation.
(1) Compared with other LMM generalists, our Lumen surpasses them by clear margins, achieving a 15.6 AP$_{50}$ boost over Griffon~\cite{zhan2023griffon} and a 4.7 AP$_{50}$ boost over InstructCV~\cite{gan2023instructcv}. This indicates the effectiveness of our method in discovering dense object cues. (2) Compared with vision generalists and task-specific specialists, our method further approaches their performances. 
We analyze that the performance gap might originate from two major aspects. First, we use the input size of $448\times448$, which is much smaller than these competitors, for example, $1024\times1024$ in~\cite{chen2021pix2seq,xu2023mplug}. Second, we do not introduce DETR-like object decoding techniques as done in~\cite{wang2023visionllm,xu2023mplug}.

\textbf{Pose Estimation.}\quad
We employ COCO human 2D body keypoint val set for evaluation.
Following the top-down paradigm employed by previous works~\cite{geng2023instructdiffusion,chen2021pix2seq}, we first crop a single
object from the image using the bounding box. Therefore, the pose estimation is simplified to discovering keypoints of a single object. Since the LMM generalists do not perform this task, we only compare our method with vision generalists and task-specific specialists. As illustrated in Table~\ref{tab:ver}, our method outperforms the vision generalist model Pix2Seq-v2~\cite{chen2021pix2seq} with 2.4 AP. Meanwhile, our Lumen also achieves comparable performances with task-specific specialists. 

\textbf{Visual Grounding \& Referring Segmentation.}\quad
Compared to object detection and instance segmentation, visual grounding and referring segmentation underscore the language comprehension ability. 
Here, we report the results on RefCOCOg val set for comparison because the referring expressions in it are more diverse and abundant than those in RefCOCO and RefCOCO+. We also provide complete results on these three benchmarks in the Appendix~\ref{appendix:vg_and_ris}. As illustrated in Table~\ref{tab:ver}, our method achieves better performances than Shikra~\cite{chen2023shikra} and Griffon~\cite{zhan2023griffon} on visual grounding task, with AP$_{50}$ 83.6 vs 82.3 and 83.6 vs 81.5, respectively. 

\textbf{Visual Question Answering.}\quad To examine general visual understanding and instruction following of our model, we follow common practices and employ MMBench~\cite{mmdetection}, MME~\cite{fu2024mme}, SEEDBench~\cite{li2023seed}, MMMU~\cite{yue2023mmmu} and MathVista~\cite{lu2024mathvista} for evaluation. As shown in Table~\ref{tab:vqa}, our Lumen achieves VQA results comparable with state-of-the-art LMMs while extending versatile vision-centric capabilities.
\begin{table}[!t]
\centering
\caption{\textbf{Ablation studies on model designs.}}
\begin{subtable}{0.3\textwidth}
    \centering
    \caption{Effect of different ``V-L dense aligner'' design choices.}
    \label{tab:abl_arch}
    \scalebox{0.8}{
    \begin{tabular}{c|ccc}
    \toprule
    Arch.              & AP   & AP$_{50}$ & AP$_{75}$ \\ \midrule
    Conv.             & 18.4 & 30.2 & 15.7 \\
    Trans.             & \textbf{28.4} & \textbf{45.0} & \textbf{28.1} \\ \bottomrule
    \end{tabular}}
    \end{subtable}
\begin{subtable}{0.3\textwidth}
    \centering
    \caption{Effect of the pretrained mask decoder. cIoUs are reported.}
    \label{tab:abl_sam}
    \scalebox{0.8}{
    \begin{tabular}{c|ccc}
    \toprule
    Mask Dec. & RC & RC+ & RCg \\ \midrule
    TransDec.  & 65.8       & 58.6        & \textbf{62.0}        \\
    SAM       & \textbf{66.6}       & \textbf{59.2}        & 61.5        \\ \bottomrule
    \end{tabular}}
\end{subtable}
\begin{subtable}{0.3\textwidth}
    \centering
    \caption{Effect of LMM baseline.}
    \label{tab:abl_lmm_baseline}
    \scalebox{0.8}{
    \begin{tabular}{c|ccc}
    \toprule
    Baseline   & AP   & AP$_{50}$ & AP$_{75}$ \\ \midrule
    LLaVA-v1.0  & 24.0 & 39.0 & 23.0 \\
    LLaVA-v1.0*  & 26.8 & 43.2 & 26.0 \\
    LLaVA-v1.5 & \textbf{28.4} & \textbf{45.0} & \textbf{28.1} \\ \bottomrule
    \end{tabular}
    }
\end{subtable}
\end{table} 
\begin{table}[!t]
\centering
\caption{\textbf{Effect of multi-task training.} AP$_{50}$ on RefCOCOg val set and COCO val set are reported.}
\label{tab:abl_task}
\scalebox{0.8}{
\begin{tabular}{c|cccc|cc}
\toprule
\# & Object Det. & Grounding & Pose Est. & VQA & RefCOCOg & COCO \\ \midrule
1  &            & \checkmark             & \checkmark         & \checkmark   & 72.3    &  26.3      \\
2  & \checkmark           &              & \checkmark         & \checkmark   & 24.6     &  41.2     \\
3  & \checkmark           & \checkmark             &          & \checkmark   & \textbf{75.8}     &  \textbf{45.7}     \\
4  & \checkmark           & \checkmark             & \checkmark         &    & 75.0   & 44.8        \\
5  & \checkmark           & \checkmark             & \checkmark         & \checkmark   & 74.8   & 45.0        \\ \bottomrule
\end{tabular}}
\end{table}

\begin{table}[!t]
\centering
\caption{\textbf{Effect of the training recipe on VQA performances.} MMBench is used for evaluation\tablefootnote{Please refer to the official benchmark for the specific meaning of each term.}.}
\label{tab:abl_training_rec}
\scalebox{0.8}{
\begin{tabular}{c|cc|cccccc|c}
\toprule
\# & Phase 1 & Phase 2 & AR            & CP            & FP-C & FP-S          & LR            & RR            & Overall       \\ \midrule
1   &     &         & \textbf{69.3} & 74.3          & 54.5 & 66.6          & 28.0          & 55.7          & 62.5          \\
2  &      & \checkmark       & 67.3          & \textbf{80.0} & 49.0 & \textbf{67.9}          & 28.8          & \textbf{58.3}          & 63.7          \\
3  &\checkmark       & \checkmark       & 67.3          & 77.0          & \textbf{58.7}  & 67.2 & \textbf{33.0} & 56.5 & \textbf{64.2} \\ \bottomrule
\end{tabular}}
\end{table}

\vspace{-8pt}
\subsection{Ablation Studies}
\label{sec:exp:comp_anl}
For efficiency, we default to training the model for 10,000 iterations in the first phase of task-agnostic learning. This protocol is standard for our ablation studies

\textbf{Multi-task Training.}\quad
We use different task combinations for training the model, and the results are shown in Table~\ref{tab:abl_task}. Based on the results, we have the following observations: (1) Compared with adopting datasets from all tasks (\#5), discarding object detection data (\#1) will reduce visual grounding performances. Similarly, removing visual grounding data (\#2) also results in performance degradation on object detection. This demonstrates that object detection and visual grounding can benefit from each other. (2) Excluding pose estimation data leads to performance enhancements on both visual grounding and object detection (\#3 vs \#5). This might be caused by the data conflict problem addressed in many generalist models~\cite{wang2023visionllm,chen2024llava}. (3) Excluding VQA data (i.e., discarding Phase 2) does not significantly affect model's performances on both detection and visual grounding, which is reasonable as VQA data do not provide explicit object-level vision-language alignment cues.

\textbf{Training Recipe.}\quad We ablate the effects of 2-phase training strategy to inspect the contribution of vision-centric data on VQA tasks. In Table~\ref{tab:abl_training_rec}, we list detailed scores on diverse aspects covered by MMBench. \#1 is initialized from LLaVA's weights, without any tuning. By comparing \#2 and \#3, we observe that FP-C (i.e, Cross-instance Perception) improves remarkably, which indicates that incorporating vision-centric data in Phase 1 can promote multi-instance perception in VQA (as well as the overall performance). Scores of Phase 1-only are not reported as the model trained without rich instruction data tends to respond template answers like \textit{``Sure, the location is }\texttt{[LOC].''}

\textbf{V-L Dense Aligner Architectures.}\quad 
We ablate different architecture design choices as shown in Table~\ref{tab:abl_arch}, where ``Conv.'' indicates the operation of concatenating feature of \texttt{[LOC]} token with every image patch features and fusing them with two convolutional layers, and ``Trans.'' represents a light-weight transformer in our main method. Substituting our transformer-based aligner with simple convolutional layers incurs significant performance degradation. This result indicates that complete vision-language interaction is crucial in promoting explicit dense vision-language alignment.

\textbf{Pretrained Mask Decoder.}\quad We adopt the mask annotations from RefCOCO (RC), RefCOCO+ (RC+) and RefCOCOg (RCg) to train a lightweight transformer-like mask decoder from scratch (TransDec.) and report cIoU on their val set to ablate the effect of leveraging pretrained SAM decoder in Table~\ref{tab:abl_sam}. Although SAM outperforms TransDec., the narrow performance gap suggests that the primary challenge in advancing segmentation performances within our framework is the need to improve the dense vision-language alignment in the learned heatmap.

\textbf{LMM Baseline.}\quad We ablate the effect of different LMM baselines on vision-centric task as shown in Table~\ref{tab:abl_lmm_baseline}. 
Since LLaVA-v1.5 employs CLIP ViT-L/14-336 as a stronger vision backbone, we use it to substitute the original CLIP ViT-L/14-224 in LLaVA-v1.0, denoted as ``LLaVA-v1.0*'' in Table~\ref{tab:abl_lmm_baseline}, to respectively ablate the effects of stronger vision backbone and multimodal language model. It can be proved that both stronger vision backbone and multimodal language model can benefit dense vision-language alignment.


\begin{figure}[!t]
  \centering
  \begin{minipage}[t]{0.45\textwidth}
    \centering
    \captionof{table}{Generalization ability evaluation on object detection. $\dagger$ indicates that VOC is not included in our training data, and thereby used for cross-dataset generalization evaluation.}
    \label{tab:generalize}
    \scalebox{0.8}{
    \begin{tabular}{l|cc}
    \toprule
    Method       & COCO          & VOC$^{\dagger}$           \\ \midrule
    RetinaNet~\cite{lin2017focal}    & -             & 77.3          \\
    Faster R-CNN~\cite{ren2015faster} & -             & 80.4          \\ \midrule
    Pix2Seq-v2~\cite{chen2021pix2seq}   & \textbf{57.4} & 38.5         \\
    InstructCV~\cite{gan2023instructcv}   & 48.5          & 61.7          \\
    Lumen (Ours) & 53.2          & \textbf{77.9} \\ \bottomrule
    \end{tabular}}
  \end{minipage}
  \hfill
  \begin{minipage}[t]{0.45\textwidth}
    \centering
     \caption{Qualitiave results of our Lumen when generalizing to the object counting task.}
     \label{Fig:counting}
    \includegraphics[width=0.9\textwidth]{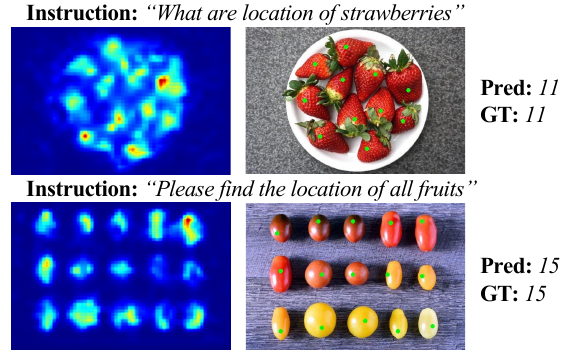}
  \end{minipage}
\end{figure}

\subsection{Generalization Evaluation}
\label{sec:exp:generalization}


\textbf{Generalization to Unseen Datasets.}\quad
To evaluate the generalization ability of our method, we perform zero-shot evaluation on PASCAL VOC2007 val set~\cite{Everingham10}. As illustrated in Table~\ref{tab:generalize}, our method demonstrates superior generalization ability than other generalist models. It is worth noting that compared with InstructCV which also inherits the enormous word vocabulary of LLM, our method outperforms it on VOC zero-shot evaluation by 16.2 AP. Besides, even compared with specialist models trained on VOC dataset (i.e, RetinaNet and Faster R-CNN in Table~\ref{tab:generalize}), our method still achieves comparable performances. 

\textbf{Generalization to Unseen Tasks.}\quad
To prove that the heatmap produced by our first-stage task-agnostic training is a powerful intermediate representation for scaling up to more vision-centric tasks, we apply simple decoding rules on the heatmap to make our Lumen support the object counting task~\cite{ranjan2021learning}. 
As shown in Fig~\ref{Fig:counting} our Lumen can make the correct prediction even without specifically training on the object counting task.

\section{Conclusions}
In this paper, we present \textbf{Lumen}, a \textbf{L}arge m\textbf{u}ltimodal \textbf{m}odel with versatile vision-centric capabilities \textbf{en}hancement. 
With the task-agnostic and task-specific design, Lumen significantly broadens the range of visual tasks that existing LMM generalists can tackle, and also maintains general visual understanding and instruction following capabilities. 


\bibliographystyle{unsrt}  
\bibliography{references}







\clearpage
\appendix

\section*{Appendix}

\section{Implementation Details}
\subsection{Detailed Design of V-L Dense Aligner}
\label{appendix:architecture}
\begin{figure}[!h]
  \centering
  \includegraphics[width=0.7\linewidth]{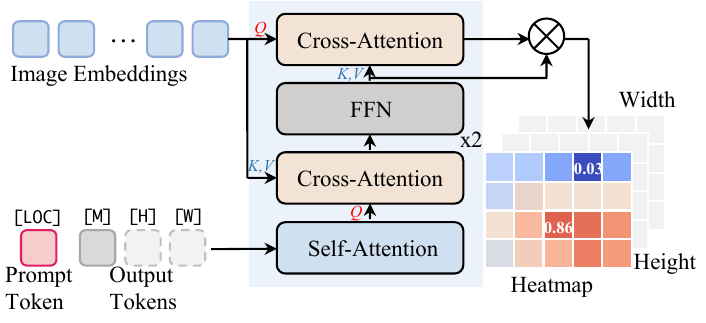}
  \caption{
  Detailed designs of ``V-L Dense Aligner''. \texttt{[LOC]} is the special token output from the LMM. \texttt{[M]} is the special token used for predicting the matching probabilities (i.e., heatmaps) in the first class-agnostic matching stage in the main paper. \texttt{[H]} and \texttt{[W]} are special tokens used as a simple box decoder for predicting additional height and width in the second class-specific decoding stage in the main paper. We use dotted lines to illustrate that these tokens are added in the second stage.
  }
  \label{fig:fig_align}
\end{figure}
The detailed design of the transformer-based V-L dense aligner is shown in Fig~\ref{fig:fig_align}. Motivated by SAM~\cite{kirillov2023segment}, we regard \texttt{[LOC]} token as a prompt token, which carries semantics conveyed in the instruction and is used to guide different output tokens, namely \texttt{[M]}, \texttt{[H]} and \texttt{[W]}, to generate corresponding predictions (i.e., heatmap, height and width in Fig~\ref{fig:fig_align}). Specifically, the prompt token and output tokens are first interacted with a self-attention layer. Then, the resulting outputs are interacted with image embeddings through a sequence of cross-attention layers and a feed-forward network as shown in Fig~\ref{fig:fig_align} to perform dual attention. The above process is repeated twice. The final outputs are produced by performing an element-wise multiplication of the features of output tokens with the features of the image embeddings. For the sake of clarity and conciseness, we have omitted channel compression steps from Fig~\ref{fig:fig_align}.

As for the output tokens, in the first stage, we only use \texttt{[M]} to generate the predicted heatmap. In the second stage, we add \texttt{[H]} and \texttt{[W]} to predict the height and width maps under the supervision of L1 loss as done in~\cite{zhou2019objects}.

\begin{table}[!t]
\centering
\caption{Performance comparison with state-of-the-art specialists and generalists on the visual grounding task.}
\label{tab:vg_supp}
\scalebox{1}{
\begin{tabular}{lcccccccc}
\toprule
\multicolumn{1}{l}{\multirow{2}{*}{Method}} & \multicolumn{3}{c}{RefCOCO}                      & \multicolumn{3}{c}{RefCOCO+}                     & \multicolumn{2}{c}{RefCOCOg}    \\
\multicolumn{1}{l}{}                        & val            & test-A         & test-B         & val            & test-A         & test-B         & val            & test           \\ \midrule
\multicolumn{9}{c}{\gray{\textbf{\textit{Specialists}}}}                                                                                                                                                      \\ \midrule
\multicolumn{1}{l}{UNINEXT~\cite{yan2023universal}}                 & \gray{92.64}          & \gray{94.33}          & \gray{91.46}          & \gray{85.24}          & \gray{89.63}          & \gray{79.79}          & \gray{88.73}          & \gray{89.73}          \\
\multicolumn{1}{l}{G-DINO-L~\cite{liu2023grounding}}                & \gray{90.56}          & \gray{93.19}          & \gray{88.24}          & \gray{82.75}          & \gray{88.95}          & \gray{75.92}          & \gray{86.13}          & \gray{87.02}          \\ \midrule
\multicolumn{9}{c}{\textbf{\textit{Generalists}}}                                                                                                                                                      \\ \midrule
\multicolumn{1}{l}{VisionLLM-H~\cite{wang2023visionllm}}             & 86.70              & -          & -              & -              & -              & -              & -              & -              \\
\multicolumn{1}{l}{OFA-L~\cite{wang2022ofa}}                   & 79.96          & 83.67          & 76.39          & 68.29          & 76.00          & 61.75          & 67.57          & 67.58          \\
\multicolumn{1}{l}{Shikra-7B~\cite{chen2023shikra}}               & 87.01          & 90.61          & 80.24          & \textbf{81.60} & 87.36 & 72.12 & 82.27          & 82.19          \\
\multicolumn{1}{l}{\textbf{Lumen-7B (Ours)}}            & \textbf{88.59} & \textbf{92.06} & \textbf{83.55} & 80.35          & \textbf{87.59}          & \textbf{72.20}          & \textbf{83.62} & \textbf{84.44} \\ \bottomrule
\end{tabular}}
\end{table}

\section{Complete Results on Visual Grounding and Referring Segmentation}
\label{appendix:vg_and_ris}
We provide the complete performance comparison with existing approaches on visual grounding and referring segmentation tasks as shown in Table~\ref{tab:vg_supp} and Table~\ref{tab:ris_supp}. We will analyze the results in the following parts.

\subsection{Visual Grounding.} We report the visual grounding performances of our Lumen on RefCOCO, RefCOCO+ and RefCOCOg datasets in Table~\ref{tab:vg_supp}. First, compared with VisionLLM-H, a generalist that can also perform object detection and instance segmentation like us, our method outperforms it on the RefCOCO val set even without using a strong vision encoder, e.g., InternImage-H used in VisionLLM-H. Besides, compared with OFA-L and Shikra-7B, which reformulate the box prediction as a sequence generation manner, our method can achieve better or comparable performances with them. It indicates that our predicted heatmap can capture the correspondence between visual regions and complex language concepts. 

\subsection{Referring Segmentation.}
We report the referring segmentation performances of our Lumen on RefCOCO, RefCOCO+ and RefCOCOg datasets in Table~\ref{tab:ris_supp}. First, compared with other generalists, our Lumen surpasses them across most benchmarks by clear margins. Besides, it is worth noting that we \textbf{do not use any pixel-level supervision} (i.e., segmentation masks) throughout our training process, but our method still achieves comparable performances with specialist models on the challenging RefCOCOg benchmark.


\begin{table*}[!h]
\centering
\caption{Performance comparison with state-of-the-art specialists and generalists on the referring image segmentation task.}
\label{tab:ris_supp}
\begin{tabular}{lccccccccc}
\toprule
\multirow{2}{*}{Method} & \multicolumn{3}{c}{RefCOCO} & \multicolumn{3}{c}{RefCOCO+} & \multicolumn{2}{c}{RefCOCOg} \\
                        & val    & test-A   & test-B  & val    & test-A   & test-B   & val           & test         \\ \midrule
\multicolumn{9}{c}{\gray{\textit{\textbf{Specialists}}}}                                                                                     \\ \midrule
MCN~\cite{luo2020multi}                     & \gray{62.4}   & \gray{64.2}     & \gray{59.7}    & \gray{50.6}   & \gray{55.0}     & \gray{44.7}     & \gray{49.2}          & \gray{49.4}         \\
LAVT~\cite{yang2022lavt}                    & \gray{72.7}   & \gray{75.8}     & \gray{68.8}    & \gray{62.1}   & \gray{68.4}     & \gray{55.1}     & \gray{61.2}          & \gray{62.1}         \\
ReLA~\cite{liu2023gres}                   & \gray{73.8}   & \gray{76.5}     & \gray{70.2}    & \gray{66.0}   & \gray{71.0}     & \gray{57.7}     & \gray{65.0}          & \gray{66.0}         \\
LISA~\cite{lai2023lisa}                    & \gray{74.1}   & \gray{76.5}     & \gray{71.1}    & \gray{62.4}   & \gray{67.4}     & \gray{56.5}     & \gray{66.4}          & \gray{68.5}         \\ \midrule
\multicolumn{9}{c}{\textit{\textbf{Generalists}}}                                                                                     \\ \midrule
X-Decoder~\cite{zou2023generalized}               & -      & -        & -       & -      & -        & -        & 64.6          & -            \\
Unified-IO~\cite{lu2022unified}              & 46.4   & 46.1     & 48.1    & 40.5   & 42.2     & 40.2     & 17.3          & -            \\
InstructDiff~\cite{geng2023instructdiffusion}            & 61.7   & 65.2     & 60.2    & 46.6   & 52.3     & 39.0     & \textbf{67.4}          & -            \\
\textbf{Lumen (Ours)}            & \textbf{70.7}   & \textbf{75.5}     & \textbf{66.2}    & \textbf{62.8}   & \textbf{69.6}     & \textbf{55.4}     & 65.1          & \textbf{67.0}         \\ \bottomrule
\end{tabular}
\end{table*}

\section{More Ablation Studies}
\subsection{Input Sizes}
\label{appendix:input_sizes}
To examine the impact of various V-L dense aligner input resolutions, we resize the input image into three distinct sizes, concurrently adjusting the positional embeddings of the original CLIP ViT. As shown in Table~\ref{tab:abl_inp}, enlarging the input size from $336\times336$ (default input size of CLIP ViT-L/14) to $448\times448$ can achieve nontrivial performance boosts. However, further increasing the size to $896\times896$ will harm the performances. This phenomenon indicates that excessively increasing the input size (to a scale dramatically different from its pretraining) will break the semantics of visual features, and thereby damaging the vision-language alignment.  

\subsection{Vision Encoders} 
\label{appendix:vision_encoders}
Since the input size can not be seamlessly scaled up as discussed above, it is worth further exploring to leverage vision encoders naturally adaptive to high-resolution image inputs. Toward this end, we employ SAM ViT-H/16~\cite{kirillov2023segment}, which takes $1024\times1024$ images as inputs, as an additional vision backbone to provide high-resolution visual features for V-L dense aligner. However, as shown in Table~\ref{tab:abl_vis}, the utilization of SAM ViT-H/16 results in a significant performance deterioration. We posit that this phenomenon can be attributed to the inadequate alignment between the SAM visual encoder and the language modality. Building upon the aforementioned experimental findings and analysis, we think that embracing vision encoders capable of processing high-resolution image inputs without compromising semantic coherence can pave the way for further enhancing the vision-centric capabilities of LMMs. 

\subsection{Number of \textit{K} for Dense Prediction Tasks}
\label{appendix:number_of_K}
We also investigate the impact of selecting different values for $K$ on the dense prediction tasks, e.g., object detection. As shown in Table~\ref{tab:abl_K}, when $K$ increases from 20 to 100, the model's performance consistently improves, which can be attributed to the increased recall of true positives. As $K$ further increases from 100 to 200, the model's performance does not improve evidently, therefore, we set $K$ as 100 in our implementation. 

\subsection{Training Epochs}
\label{appendix:training_epochs}
We record the variations in detection performance as the number of training iterations increases in Table~\ref{tab:abl_iter}. It can be observed that the model's performance slowly increases as the training proceeds.
Due to the limitation of computational resources, we did not train our model for a longer time.
The convergence speed might be limited by the optimization difficulty of utilizing a single \texttt{[LOC]} token to query the entire image regions. A feasible solution to mitigate this problem is to employ more special tokens like \texttt{[LOC]} serving as object queries in DETR~\cite{carion2020end} with the Hungarian matching for label assignment, which can accelerate training and promote performances as proved in VisionLLM~\cite{wang2023visionllm}. We do not adopt this design as it is unclear whether these tokens customized for object detection can generalize well to other tasks.

\begin{table}[!t]
\centering
\caption{\textbf{Ablation studies on model designs.}}
\begin{subtable}{0.4\textwidth}
    \centering
    \caption{Effect of different input sizes for the dense alignment.}
    \label{tab:abl_inp}
    \scalebox{0.9}{
    \begin{tabular}{c|ccc}
        \toprule
        Input Size & AP            & AP$_{50}$          & AP$_{75}$          \\ \midrule
        336$\times$336    & 26.1          & 40.2          & 25.6          \\
        448$\times$448    & \textbf{28.4} & \textbf{45.0} & \textbf{28.1} \\
        896$\times$896    & 24.5          & 36.4          & 22.5          \\ \bottomrule
    \end{tabular}}
\end{subtable}
\begin{subtable}{0.5\textwidth}
    \centering
    \caption{Effect of different pre-trained vision encoders.}
    \label{tab:abl_vis}
    \scalebox{0.9}{
    \begin{tabular}{c|ccc}
    \toprule
    Vis Enc.   & AP   & AP$_{50}$ & AP$_{75}$ \\ \midrule
    SAM ViT-H/16-1024  & 14.5 & 24.0 & 14.1 \\
    CLIP ViT-L/14-336 & \textbf{28.4} & \textbf{45.0} & \textbf{28.1} \\ \bottomrule
    \end{tabular}
    }
\end{subtable}
\begin{subtable}{0.45\textwidth}
    \centering
    \caption{Effect of different $K$ value on dense prediction task.}
    \label{tab:abl_K}
    \scalebox{0.9}{
    \begin{tabular}{c|cccc}
        \toprule
        $\#K$ & 20 & 50 & 100  & 200 \\ \midrule
        AP   & 26.3  & 27.6  & 28.4 & \textbf{28.5}   \\
        AP$_{50}$ & 39.5  & 43.2  & 45.0 & \textbf{45.3}   \\
        AP$_{75}$ & 25.3  & 27.2  & \textbf{28.1} & 28.0   \\ \bottomrule
    \end{tabular}}
\end{subtable}
\begin{subtable}{0.45\textwidth}
    \centering
    \caption{Effect of different numbers of training iterations.}
    \label{tab:abl_iter}
    \scalebox{0.9}{
    \begin{tabular}{c|ccccc}
    \toprule
    \#Iteration & 10k  & 20k & 30k & 40k & 50k  \\ \midrule
    AP    & 28.4 & 31.6   & 34.5   & 35.2   & \textbf{35.3} \\
    AP$_{50}$  & 45.0 & 49.2   & 52.4   & 53.2   & \textbf{53.2} \\
    AP$_{75}$  & 28.1 & 31.5   & 34.7   & 35.5   & \textbf{35.8} \\ \bottomrule
    \end{tabular}
    }
\end{subtable}
\end{table}


\begin{figure}[!ht]
  \centering
  \includegraphics[width=\linewidth]{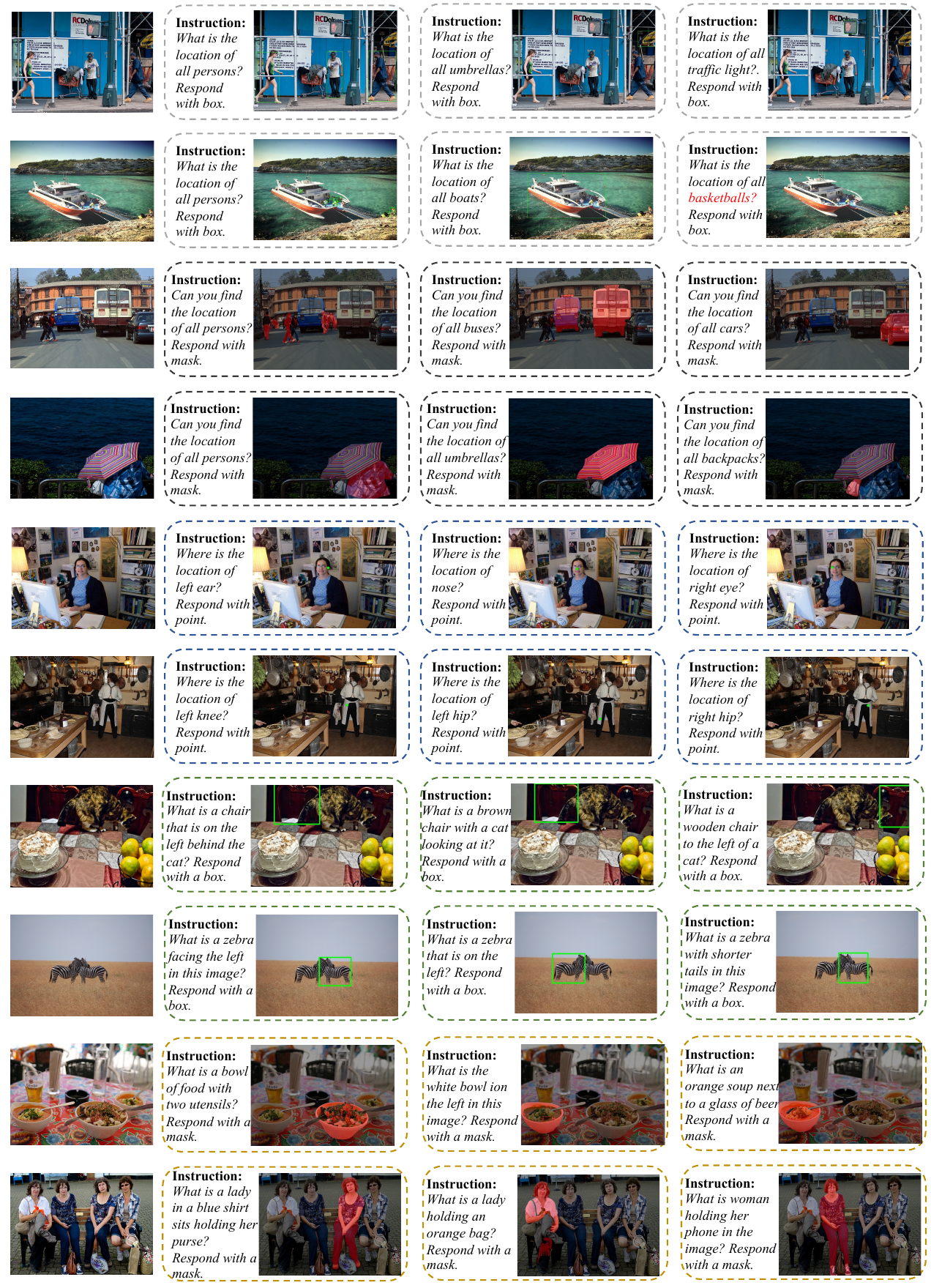}
  \caption{
  Qualitative results of versatile vision-centric capabilities of our proposed Lumen. We use different colors of dotted rectangles to denote performing different tasks.
  }
  \label{fig:fig_qual}
\end{figure}

\section{More Qualitative Results}
We provide qualitative results to demonstrate the versatile capabilities of our Lumen as shown in Fig~\ref{fig:fig_qual}. Overall, the abundant qualitative results prove that our method possesses versatile vision-centric capabilities. It is worth noting that in the case within the second row, we instruct the model to detect the non-existing object (i.e., \textit{``basketball''} in Fig~\ref{fig:fig_qual}), the maximum activation across the heatmap is relatively low, and filtered with a pre-defined threshold. Therefore, the model will not generate any output for this wrong instruction as illustrated in Fig~\ref{fig:fig_qual}.


\end{document}